\documentclass[runningheads]{llncs}

\usepackage[mobile]{eccv}

\usepackage{eccvabbrv}
\usepackage{graphicx}
\usepackage{booktabs}
\usepackage{amsmath}

\usepackage{subcaption}
\usepackage{multirow}
\usepackage{url}
\usepackage{wrapfig}
\usepackage{xcolor}
\usepackage{lineno}
\usepackage{geometry}

\usepackage[accsupp]{axessibility}

\usepackage{hyperref}
\usepackage{orcidlink}

\newcommand{\Figure}[1]{Fig.~\ref{#1}}
\newcommand{\Table}[1]{Table~\ref{#1}}
\newcommand{\Section}[1]{Sec.~\ref{#1}}

\begin{document}

\title{Monitoring Pasture Restoration from Satellite Image Time Series: Caveats and Opportunities}
\titlerunning{Monitoring Pasture Restoration from SITS}

\author{
Linnea Sartorius\inst{1}\thanks{Equal contribution. Work conducted while at RISE Research Institutes of Sweden.} \and
Isak Randahl\inst{1\star} \and
Delia Fano Yela\inst{2,3} \and
Georg Andersson\inst{2} \and
Sadegh Jamali\inst{1} \and
Aleksis Pirinen\inst{2,3,4}
}

\authorrunning{L. Sartorius et al.}

%\institute{
%$^1$ Lund University \hspace{6pt} $^2$ RISE Research Institutes of %Sweden \hspace{6pt} $^3$ Climate AI Nordics
%}
\institute{Lund University \and RISE Research Institutes of Sweden \and Climate AI Nordics \and Swedish Centre for Impacts of Climate Extremes (CLIMES)}

\maketitle

\begin{abstract}
Monitoring nature restoration at scale is an important but difficult ecological problem. Deep learning methods to analyze satellite image time series (SITS) have been widely used for land surface monitoring. In semi-natural grasslands -- the habitat type in focus in this work -- restoration outcomes develop gradually, yet satellite observations are influenced by weather, acquisition conditions, and processing artefacts, making it difficult to distinguish genuine restoration signals from unrelated temporal variation. In this work, we examine -- to the best of our knowledge, for the first time -- whether restoration status can be detected directly from satellite image time series by formulating pasture restoration as a binary deep learning classification problem. We evaluate two common SITS deep learning architectures on different Sentinel-2 image combinations, across 1,397 restored Swedish pastures and find that explicitly modeling intra-year variability and per-pasture normalization increases separability, reaching 0.88 accuracy for the best model. We further investigate our results and perform a targeted bias analysis finding that reliable deployment requires temporally balanced labels and evaluation protocols that explicitly test for year-related confounding. We therefore frame our contribution not as a solved restoration-monitoring system, but as a realistic case study of what works, what fails, and what future studies should control for. Code and models are available at \url{https://github.com/aleksispi/ml-nature-resto}.

\end{abstract}

\section{Introduction}\label{sec:intro}
Nature restoration has become a policy priority across Europe, including the recovery of degraded grassland ecosystems that support biodiversity, pollination, and resilient agricultural landscapes \cite{eu2024,naturvardsverket2024}. In Sweden, semi-natural pastures are ecologically valuable but have declined substantially as small-scale grazing systems have been abandoned \cite{lindborg2004,ke2020}. Monitoring restoration progress in these landscapes is difficult in practice: field inspections are informative but costly, infrequent, and difficult to scale to national programs. Satellite image time series (SITS) offer an attractive complementary signal as they provide repeated observations over large areas \cite{phiri2020,miller2024}. For example, SITS have recently been used to monitor grazing activity in semi-natural pastures on a season-by-season basis \cite{pirinen2025grazing}. However, nature restoration poses a distinct problem: rather than detecting management activity within a single growing season, the objective is to characterize ecological recovery trajectories that unfold over several years.

Unlike abrupt land-cover change, restoration is gradual, heterogeneous, and only partially observable from reflectance trajectories. The ecological process unfolds over multiple years, while satellite observations are shaped by seasonality, cloud cover, geography, and sensor-processing changes \cite{forkel2013,clearsky2026}. In other words, the same modeling machinery that can detect restoration-related change can also exploit time-dependent artifacts that happen to correlate with restoration labels. This issue is particularly acute when restoration labels are strongly correlated with calendar year, for example as in this work, when early years predominantly correspond to non-restored states and later years to restored states.

In this paper, we examine these issues in the specific context of pasture-level restoration monitoring using Sentinel-2 summer time series over Sweden.
\begin{wrapfigure}{r}{0.34\linewidth} % r = right placement
    \vspace{-9pt}
    \centering
    \includegraphics[width=\linewidth]{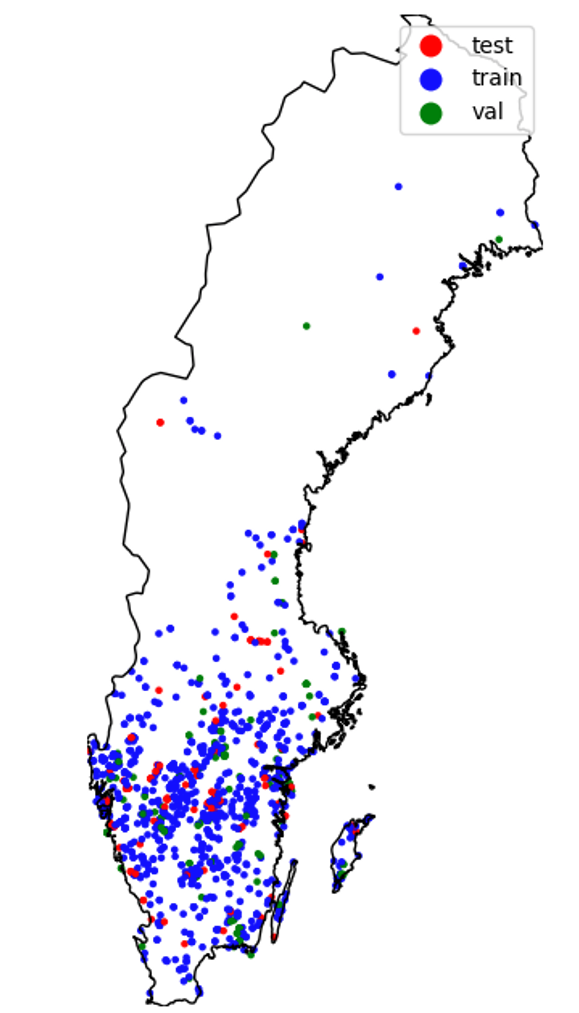}
    \caption{Spatial distribution of training, validation, and test polygons across Sweden.}
    \vspace{-9pt}
    \label{fig:sweden-polys}
\end{wrapfigure}
We formulate whether a pasture has been restored or not -- to the best of our knowledge, for the first time -- as a binary classification deep learning task, 
and evaluate two common architectures identified by \cite{miller2024} on different input representations. More precisely, we compare 2D CNNs operating on yearly composites and monthly composite stacks against CNN--LSTM sequence models, and we evaluate a standard global z-score normalization against a more adaptive per-pasture normalization. We further examine the influence of input features, the robustness of key modeling assumptions, and the experimental controls required to support credible claims of restoration detection.

Our results show that detecting nature restoration is feasible, and that both temporal information and normalization strategy influence classification performance. However, our bias analyses show that good restoration accuracy alone is insufficient evidence that the model has learned restoration signals. The dataset contains strong year-related signatures, and models can classify calendar year surprisingly well even when the restoration target is masked or weakened. We therefore also propose a reduced-bias data representation designed to suppress year-specific cues.

In summary, we \textbf{(i)} tackle pasture-level nature restoration monitoring using SITS, and compare composite-based CNNs and sequence-based CNN--LSTM models under several input representations; \textbf{(ii)} show that leveraging intra-year variability and per-pasture normalization improves performance, while revealing substantial temporal confounding in the data; \textbf{(iii)} distill practical lessons for computer vision for ecology: strong benchmark performance is possible on real restoration data, but reliable deployment requires temporally balanced data and bias-oriented evaluation.

\section{Dataset Overview}\label{sec:dataset}
In this work we leverage pasture polygons
and restoration metadata provided by the Swedish Board of Agriculture together with Sentinel-2 L2A imagery acquired between 2018 and 2025. The data contains pastures that received restoration support, including polygon geometry, restoration start year, and the year where restoration was approved. We define the \emph{last year} as the summer season one year \emph{after} approval to ensure a complete restored season is captured. After filtering non-polygon artifacts, splitting multi-polygons, and removing very small areas ($\leq 400\,\mathrm{m}^2$, i.e.~$2\times 2$ pixels), the final dataset contains 1,397 pasture polygons. See
\Figure{fig:sweden-polys} for the spatial distribution across Sweden.

\begin{figure}[t]
    %\vspace{-6pt}
    \centering
        \begin{subfigure}[b]{0.325\linewidth}
            \includegraphics[width=\linewidth]{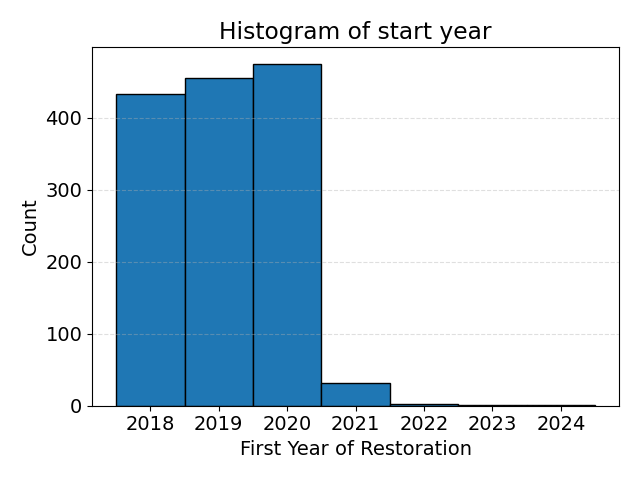}
            %\caption{Histogram showing spread of start years for the pastures}
        \end{subfigure}
        \hfill
        \begin{subfigure}[b]{0.325\linewidth}
            \includegraphics[width=\linewidth]{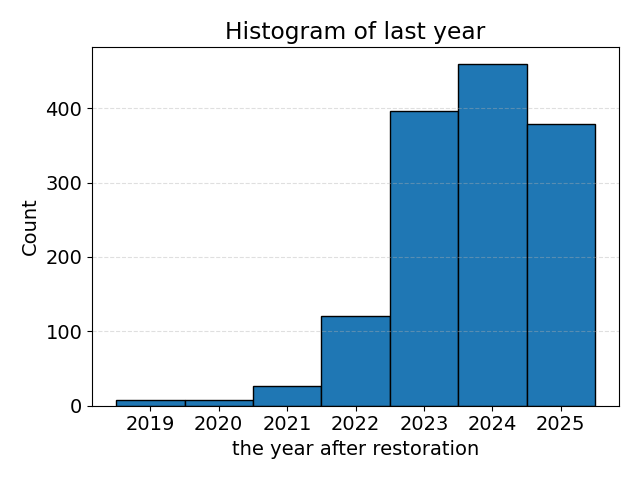}
            %\caption{Histogram showing spread of last years for the pastures}
        \end{subfigure}
        \hfill
        \begin{subfigure}[b]{0.325\linewidth}
            \includegraphics[width=\linewidth]{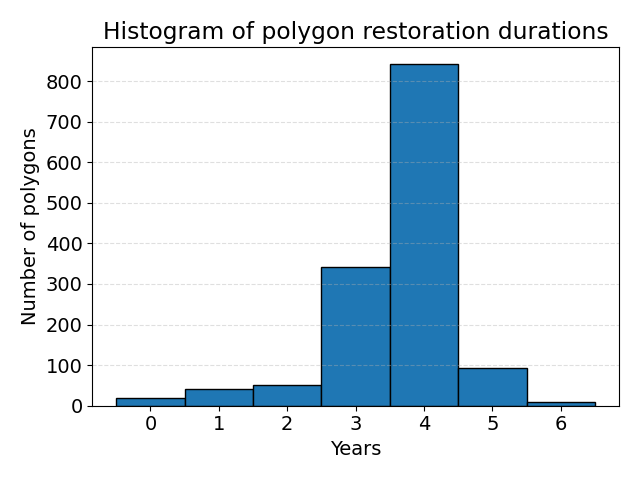}
            %\caption{TODO. Some text some text some text some text}
        \end{subfigure}
        %\vspace{-5pt}
        \caption{Histograms of the spread of start of restoration (left) and end of restoration (middle) years for the 1,397 field polygons, and the restoration duration extents (right).}
        \label{fig:resyears}
        %\vspace{-7pt}
\end{figure}

For each polygon we use Sentinel-2 summer observations, June to August. Bands at 20\,m and 60\,m resolution are bilinearly upsampled to 10\,m and stored in a common tensor representation padded to $100\times100$ pixels. The upsampling is performed solely to place all spectral bands on a common spatial grid for subsequent CNN processing; it does not increase the effective spatial resolution or introduce new spatial information. Since the interpolated bands retain their original information content, the network can only exploit spatial detail present at the native Sentinel-2 resolutions. Pixels outside the target polygon are masked to suppress context from nearby fields and farms. This is important because multiple pastures can occur within the same satellite tile and, without masking, nearby land-use dynamics can leak into the prediction task.

Cloud filtering is performed using a learned cloud detector \cite{pirinen2024creating}. Importantly, the preprocessing accounts for the 2022 Sentinel-2 processing-baseline offset: filtering with the radiometric offset applied introduced a visible temporal trend in image availability, so the final pipeline 
\begin{wrapfigure}{r}{0.46\linewidth} % r = right placement
    \vspace{-10pt}
    \centering
    \includegraphics[width=\linewidth]{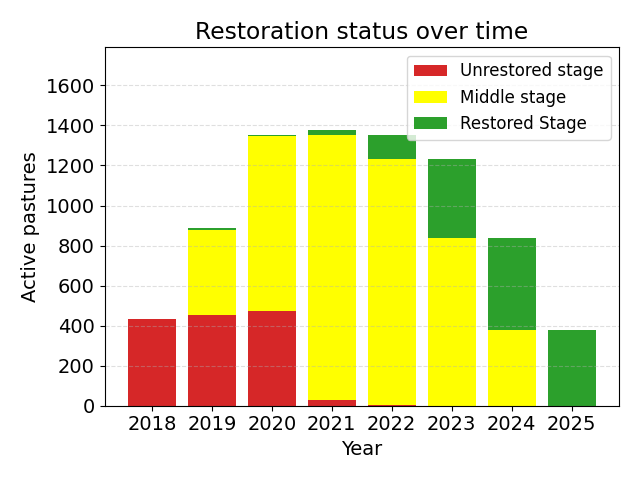}
    \vspace{-14pt}
    \caption{Restoration statuses across years.}
    \label{fig:status-hist}
    \vspace{-14pt}
\end{wrapfigure}
filters clouds without the offset in order to reduce additional year-related bias. Images with missing values, negative reflectance anomalies, or clear channel-wise outlier statistics are removed. After filtering, approximately 39\% of the original images remain, corresponding to roughly between 11 and 18 valid summer images per polygon-year depending on the year.

The classification target is derived from the restoration timeline (see \Figure{fig:resyears}). For the main binary setup, the first available year for a pasture is labeled \emph{non-restored} and the last year is labeled \emph{restored}. This gives two samples per pasture and therefore a balanced binary dataset by construction. The middle years are retained for bias analysis and for sequence-model experiments, but their labels are intrinsically more ambiguous because the degree of restoration is not directly observed for each intermediate year (see also \Figure{fig:status-hist}). A critical property of the dataset is that class label and calendar year are strongly entangled: early years are dominated by non-restored samples and late years by restored samples, introducing an unavoidable bias. To avoid spatial leakage, train/validation/test splits are made by clustering nearby polygons with a 1.5\,km buffer and assigning entire clusters to the same split, while preserving key geographical variability such as growth zones and precipitation regimes.  The resulting split is roughly 78.7\% / 12.4\% / 8.9\% in polygons for training, validation, and test.

\section{Methods}\label{sec:method}
There are four principal families of deep learning architectures for SITS analysis identified by  \cite{miller2024}: recurrent neural networks (RNNs) such as Long Short-Term Memory (LSTM)~\cite{hochreiter1997lstm}, convolutional neural networks (CNNs), hybrid CNN-RNN models, and transformer models. Given our task and data constraints, here we explore CNNs and hybrid CNN-LSTM architectures.  
\\ \\
\textbf{2D CNNs on seasonal composites.}
The input to the 2D CNN is a yearly seasonal composite formed by taking the pixel-wise median over all valid summer images for one pasture-year. We compare this single-composite representation to
%three
two alternatives:
%(i) a random single image sampled from the year,
(i) a composite augmented with vegetation indices (NDVI, EVI, NDMI, and MSAVI), and (ii) a \emph{composite stack} in which June, July, and August are each summarized separately and stacked into a 36-channel input. The composite stack is intended to preserve intra-year variation without requiring full recurrent modeling.
Two CNN backbones are evaluated: (i) a ResNet50 \cite{he2016deep}, and (ii) a lighter custom residual CNN tailored to the dataset size. The custom model consists of an initial convolutional layer, two residual blocks at 32 channels, downsampling to 64 channels, two further residual blocks, global average pooling, and a small multilayer perceptron classifier. Models are trained with Adam \cite{kingma2014adam}, binary cross-entropy with logits, random flips and $90^\circ$ rotations, and early stopping using validation performance.
\\ \\
%\textbf{CNN--LSTM sequence models.} To model sequential information directly, we use a CNN--LSTM architecture that extracts per-image spatial descriptors with the convolutional stem of the custom CNN and maps the resulting feature vectors through a one-layer LSTM \cite{hochreiter1997long}. An FC-layer operating on the final hidden state then yields a decision.
\textbf{CNN--LSTM sequence models.} To model sequential information directly, we extract per-image spatial descriptors with the custom CNN and map the resulting feature vectors through a one-layer LSTM \cite{hochreiter1997long}. An FC-layer operating on the final hidden state then yields a decision.
%We evaluate two readout strategies: the final hidden state and an attention-weighted sum over time.
Sequence inputs are built in two ways: (i) all valid images from the first year vs. all valid images from the last year, and (ii) multi-year sequences where the early half of the restoration trajectory is treated as negative and the late half as positive. The latter leverages a larger portion of the restoration trajectory, but also introduces additional label ambiguity (cf.~\Figure{fig:status-hist}).

\subsection{Normalization Schemes and Evaluation Analyses}\label{sec:norm-and-bias}
We compare two normalization schemes: (i) \emph{global normalization} uses training-set means and standard deviations for each band; (ii) \emph{per-pasture normalization} instead standardizes each pasture using statistics aggregated over its own restoration period. The latter is motivated by the possibility that restoration manifests more consistently as a within-pasture shift than as an absolute spectral state shared across different landscapes.

Because restoration labels are strongly entangled with calendar year, standard accuracy metrics alone are insufficient to assess what the models learn. We therefore complement restoration classification with two complementary analyses. First, a \emph{restoration trajectory analysis} evaluates how often the model predicts ``restored'' at different years of a restoration trajectory, which provides insights on how predictions evolve over time (see \Figure{fig:when_res_pasture} in \Section{sec:results}). 
%Second, a \emph{pasture-normalization bias} test simulates partial trajectories to determine whether per-pasture normalization leaks future information.
Second, \emph{temporal bias} is assessed in \Section{sec:res-temporal-bias} by training models to classify calendar year itself, both on regular masked inputs and on \emph{reverse-masked} inputs where the pasture is removed and only the surrounding scene remains.
%If year can be predicted well in this setting, then restoration performance must be interpreted cautiously.

\section{Experimental Results}\label{sec:results}
\Table{tab:mainresults} shows the key quantitative findings. The strongest non-sequential (Custom CNN) representation is the monthly composite stack. With global normalization, the composite stack reaches 0.75 accuracy, improving over the single seasonal composite (0.73). With per-pasture normalization, performance increases substantially to 0.86 accuracy, with balanced specificity (0.88) and recall (0.84). This strongly suggests that intra-year variability is useful and that restoration is easier to detect as a within-pasture change than as a universal end-state shared across sites. We also see (on validation data; bottom two rows of \Table{tab:mainresults}) the custom CNN architecture outperforms ResNet50.

\begin{table}[t]
%\vspace{-6pt}
\centering
\caption{Main results (on the test set unless otherwise specified). Monthly composite stacks outperform a single seasonal composite, and per-pasture normalization yields a substantial performance gain. Sequential CNN--LSTM models provide a further slight improvement, with the multi-year variant achieving the best overall accuracy (0.88) and recall (0.92).}
%\vspace{-2pt}
\label{tab:mainresults}
\scalebox{0.97}{
\begin{tabular}{lccc}
\toprule
Model / input & Accuracy & Specificity & Recall \\
\midrule
Custom CNN, single composite & 0.73 & 0.70 & 0.75 \\
Custom CNN, monthly composites stack & 0.75 & 0.67 & 0.83 \\
Custom CNN, monthly composites, added veg indices & 0.72 & 0.62 & 0.81 \\
Custom CNN, monthly composites, per-pasture norm & 0.86 & \textbf{0.88} & 0.84 \\
\midrule
CNN--LSTM, first/last year, per-pasture norm & 0.85 & 0.84 & 0.88 \\
CNN--LSTM, multi-year, per-pasture norm & \textbf{0.88} & 0.85 & \textbf{0.92} \\
\midrule
Custom CNN, 3-image random stack + per-pasture norm & 0.76 & 0.73 & 0.79 \\
\midrule
Custom CNN, monthly composites, per-pasture norm (\textbf{val})  & 0.88 & 0.89 & 0.88 \\
ResNet50, monthly composites, per-pasture norm (\textbf{val}) & 0.80 & 0.78 & 0.81 \\
\bottomrule
\end{tabular}}
%\vspace{-10pt}
\end{table}

\begin{figure}[t]
    \centering
        \begin{subfigure}[b]{0.43\linewidth}
            \includegraphics[width=\linewidth]{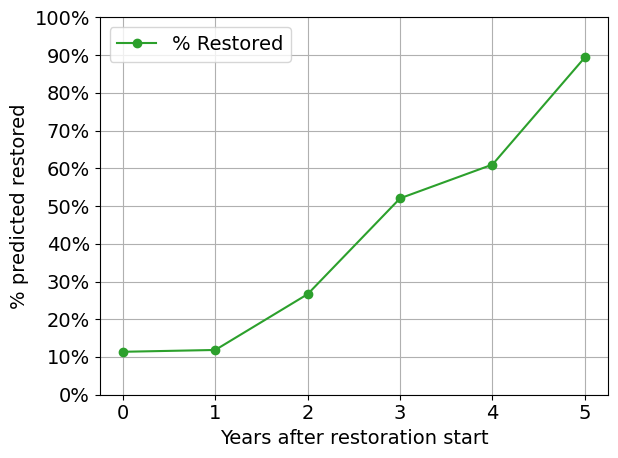}
            %\caption{Trial done for the validation set}
        \end{subfigure}
        \hfill
        \begin{subfigure}[b]{0.43\linewidth}
            \includegraphics[width=\linewidth]{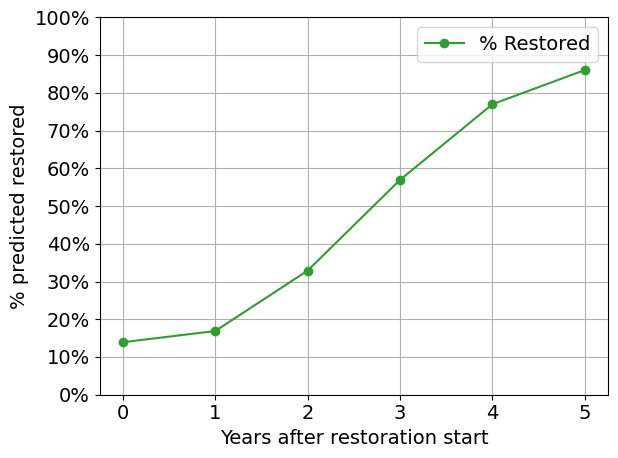}
            %\caption{Trial done for the test set}
        \end{subfigure}
        %\vspace{-4pt}
        \caption{Restoration trajectory analysis (left: val, right: test) on the pasture normalized 2D CNN model. The model predicts very few samples as \emph{restored} in the early years and almost 90\% of samples as \emph{restored} in the final year.}
        \label{fig:when_res_pasture}
        %\vspace{-10pt}
\end{figure}

The sequential (CNN--LSTM) models are competitive and slightly outperform the best 2D CNN. The best sequence model uses multi-year input and per-pasture normalization, with 0.88 accuracy, 0.85 specificity, and 0.92 recall. This indicates that sequence models can exploit temporal information effectively, especially for identifying restored samples. The improvement over the simpler first/last-year variant suggests that modeling longer temporal context provides additional signal beyond a single early–late comparison. However, the modest gain over the composite-stack CNN is small, suggesting that monthly summaries capture a large portion of the useful signal.
\\ \\
\textbf{Restoration trajectory behavior.} The restoration trajectory analysis (\Figure{fig:when_res_pasture}), applied to the best-performing 2D CNN (monthly composite stack with per-pasture normalization), shows that the model assigns progressively higher restoration probabilities to later years of the trajectory. Predicted restoration probabilities increase monotonically along the restoration trajectory, from low values in the early years to high values in the final years. This indicates that the model captures a temporal ordering signal consistent with restoration progression. However, because calendar year and restoration status are strongly correlated in the dataset (cf.~\Figure{fig:status-hist}), this monotonic trend may arise either from genuine restoration signals or from the model implicitly predicting the acquisition year. The trajectory alone therefore does not distinguish between these explanations.

\subsection{Investigating temporal confounding}\label{sec:res-temporal-bias}
Table~\ref{tab:anti-mask} shows that calendar year can be predicted with high accuracy from the data, confirming the presence of strong temporal confounding. Using reverse-masked inputs that exclude the pasture entirely, the model achieves an F1 score of 0.72, which indicates that substantial year-specific signal exists in the surrounding context. Even when restricting inputs to the pasture region, year prediction remains well above chance (F1-score 0.49 vs.\ 0.125), which demonstrates that temporal information is also encoded within the fields themselves. Per-pasture normalization reduces year predictability in the reverse-masked setting (0.72→0.64) but increases it in the masked setting (0.49→0.56), which suggests a non-trivial interaction with within-pasture temporal structure. These results show that temporal bias operates at both global and local levels, and cannot be eliminated by masking alone.

\begin{table}[b]
\centering
\caption{Year-prediction results on the validation set. All results are based on using a Custom 2D CNN, and except otherwise specified a single composite per year (the last two rows which use a random image from each of June, July, August), respectively.}
\vspace{4pt}
\label{tab:anti-mask}
\scalebox{1.05}{
\begin{tabular}{lc}
\toprule
Model / input & F1-macro  \\
\midrule
Reverse masking & 0.72 \\
Reverse masking + per-pasture norm & 0.64 \\
Regular masking & 0.49 \\
Regular masking + per-pasture norm & 0.56 \\
\midrule
Reverse masking, 3 random image stack (run 1) & 0.24 \\
Reverse masking, 3 random image stack (run 2) & 0.30 \\
\bottomrule
\end{tabular}}
%\vspace{-8pt}
\end{table}

\vspace{9pt}
\noindent\textbf{Reducing temporal shortcuts and residual signal.} To better understand the extent to which models rely on temporal shortcuts, we construct a reduced-bias variant that lowers cloud-filter thresholds and replaces monthly composites with a randomly sampled three-image stack per epoch (one image each from June, July, August). This weakens consistent seasonal structure and reduces the predictability of calendar year (F1 $\approx$ 0.24--0.30). If restoration performance were entirely driven by year prediction, this reduction should lead to a collapse in accuracy. Instead, restoration classification remains reasonably strong, with a test accuracy of 0.76 (third-to-last row in Table~\ref{tab:mainresults}), compared to 0.86 for the standard per-pasture normalized composite stack. This drop is meaningful but moderate, which indicates that temporal confounding contributes to performance, while the relatively high remaining accuracy suggests the presence of a non-trivial restoration-related signal that is being detected by the model.

%\paragraph{Summary of results and practical interpretation.}
%Taken together, the experiments support three conclusions. First, temporal structure within a growing season is informative and should not be collapsed too aggressively. Second, expensive sequence models are not always necessary; a carefully designed composite-stack CNN can be competitive. Third, and most importantly, realistic ecological ML studies need explicit tests for confounding. Reporting only restoration accuracy would considerably overstate what the present data can support.

\section{Conclusions}\label{sec:conclusions}
\iffalse
    In this work we have presented a deep learning approach to examine whether pasture-level nature restoration can be monitored from Sentinel-2 time series. The answer is cautiously positive as, although both the CNN and CNN–LSTM models achieve strong discrimination, an unavoidable temporal confound remains, preventing us from attributing the results solely to the restoration process. We propose a solution to alleviate such a temporal bias, that despite reducing the classification accuracy from 0.86 to 0.76, it improves model robustness. 
    
    Our analysis shows that within-season variability matters, that per-pasture normalization can be powerful in heterogeneous grassland settings, and that targeted temporal bias tests are essential for credible interpretation. Most importantly, we demonstrate that strong benchmark metrics alone do not constitute sufficient evidence of robust ecological monitoring.
    
    Future datasets integrated in our proposed workflow should contain temporally balanced examples in which restored and non-restored states coexist within the same years, as well as truly negative trajectories that never become restored. Until then, SITS deep learning methods for nature restoration monitoring should be viewed as promising yet methodologically constrained, capable of capturing ecologically relevant signals but highly sensitive to confounding unless study design is explicitly built around it.
\fi
In this work we have investigated whether pasture-level nature restoration can be monitored from Sentinel-2 satellite image time series (SITS) using deep learning. Our findings provide a cautiously positive answer. Both CNN and CNN--LSTM models achieve strong discrimination between restored and non-restored pasture states, which indicates that SITS contain information relevant to restoration monitoring. However, our analyses reveal substantial temporal confounding in the dataset, which prevents the observed performance from being attributed exclusively to restoration-related signals. To address this issue, we introduced a reduced-bias evaluation setting designed to weaken year-specific shortcuts. Although this reduced classification accuracy from 0.86 to 0.76, the remaining performance suggests that meaningful restoration-related information is present and leveraged by the model.
%while improving the robustness of the evaluation.

Our analysis shows that within-season variability is informative, that per-pasture normalization can be particularly effective in heterogeneous grassland environments, and that targeted temporal bias analyses are essential. Most importantly, we demonstrate that strong benchmark performance alone does not provide sufficient evidence of robust ecological monitoring.

Future datasets integrated with this workflow should contain temporally balanced examples where restored and non-restored states coexist within the same years, as well as genuinely negative trajectories that never become restored. Until such data become available, SITS-based deep learning methods for nature restoration monitoring should be viewed as promising but methodologically constrained: capable of capturing ecologically relevant signals, yet highly sensitive to temporal confounding unless study design and evaluation protocols are explicitly constructed to control for it. Code and models are available at \url{https://github.com/aleksispi/ml-nature-resto}.

\section*{Acknowledgments}
This work was funded by the Swedish National Space Agency (project number 2025-00013). We are grateful for the support and data provided by the Swedish Board of Agriculture, and Niklas Boke Ol\'en in particular.

%\clearpage

% ---- Bibliography ----
\bibliographystyle{splncs04}
\bibliography{bibliography}

\end{document}